\def\year{2020}\relax
\documentclass[letterpaper]{article} % DO NOT CHANGE THIS
\usepackage{aaai20}  % DO NOT CHANGE THIS
\usepackage{times}  % DO NOT CHANGE THIS
\usepackage{helvet} % DO NOT CHANGE THIS
\usepackage{courier}  % DO NOT CHANGE THIS
\usepackage[hyphens]{url}  % DO NOT CHANGE THIS
\usepackage{graphicx} % DO NOT CHANGE THIS
\usepackage{graphicx}  % DO NOT CHANGE THIS
\usepackage{multirow}
\usepackage{subcaption}
\usepackage{color}
\usepackage{amssymb}
\usepackage{gensymb}
\usepackage{enumitem}
\usepackage[linesnumbered, ruled]{algorithm2e}
\usepackage{amsmath}

\newcommand{\ours}{{LGnet}\xspace}

\author{Xianfeng Tang$^\dagger$, Huaxiu Yao$^\dagger$, Yiwei Sun$^\dagger$, Charu Aggarwal$^\ddagger$, Prasenjit Mitra$^\dagger$, Suhang Wang$^\dagger$\thanks{Corresponding Author}\\
$^\dagger$College of Information Sciences and Technology, Pennsylvania State University, PA, USA\\
$^\ddagger$IBM T.J. Watson, NY, USA\\
$^\dagger$\{xut10,huy144,yus162,pum10,szw494\}@psu.edu; $^\ddagger$charu@us.ibm.com
}

\title{Joint Modeling of Local and Global Temporal Dynamics for \\ Multivariate Time Series Forecasting with Missing Values}

\begin{document}

\maketitle

\begin{abstract}
Multivariate time series (MTS) forecasting is widely used in various domains, such as meteorology and traffic.
Due to limitations on data collection, transmission, and storage, real-world MTS data usually contains missing values, making it infeasible to apply existing MTS forecasting models such as linear regression and recurrent neural networks.
Though many efforts have been devoted to this problem, most of them solely rely on local dependencies for imputing missing values, which ignores global temporal dynamics. Local dependencies/patterns would become less useful when the missing ratio is high, or the data have consecutive missing values; while exploring global patterns can alleviate such problem. Thus, jointly modeling \textit{local} and \textit{global} temporal dynamics is very promising for MTS forecasting with missing values. However, work in this direction is rather limited. 
Therefore, we study a novel problem of MTS forecasting with missing values by jointly exploring local and global temporal dynamics. We propose a new framework \ours, which leverages memory network to explore global patterns given estimations from local perspectives. We further introduce adversarial training to enhance the modeling of global temporal distribution. 
Experimental results on real-world datasets show the effectiveness of \ours for MTS forecasting with missing values and its robustness under various missing ratios.
\end{abstract}

\section{Introduction}
Multivariate time series (MTS) forecasting is widely used in many applications such as weather forecasting \cite{xingjian2015convolutional}, clinical diagnosis \cite{che2018recurrent}, sales forecasting \cite{wu2018restful,wu2019neural} and traffic analysis \cite{yao2019revisiting,yao2018deep,yao2019learning,tang2019joint}. 
The popularity of MTS forecasting has attracted increasing attention, and many efforts have been taken to address the problem in the past few years~\cite{box2015time,qin2017dual,chang2018memory}.
Recurrent neural networks (RNNs), a class of deep learning frameworks designed for modeling sequential data, 
have been successfully applied to this problem. For example, 
LSTNet ~\cite{lai2018modeling} adopts LSTM to capture long-term dependencies for time series forecasting.
\citeauthor{qin2017dual}~\cite{qin2017dual} designed two attention mechanisms in RNN to improve forecasting accuracy.
Despite their success, \textit{the majority of the aforementioned models assume MTS data is complete.}

\begin{figure}[t]
\begin{center}
    \includegraphics[width=.70\columnwidth]{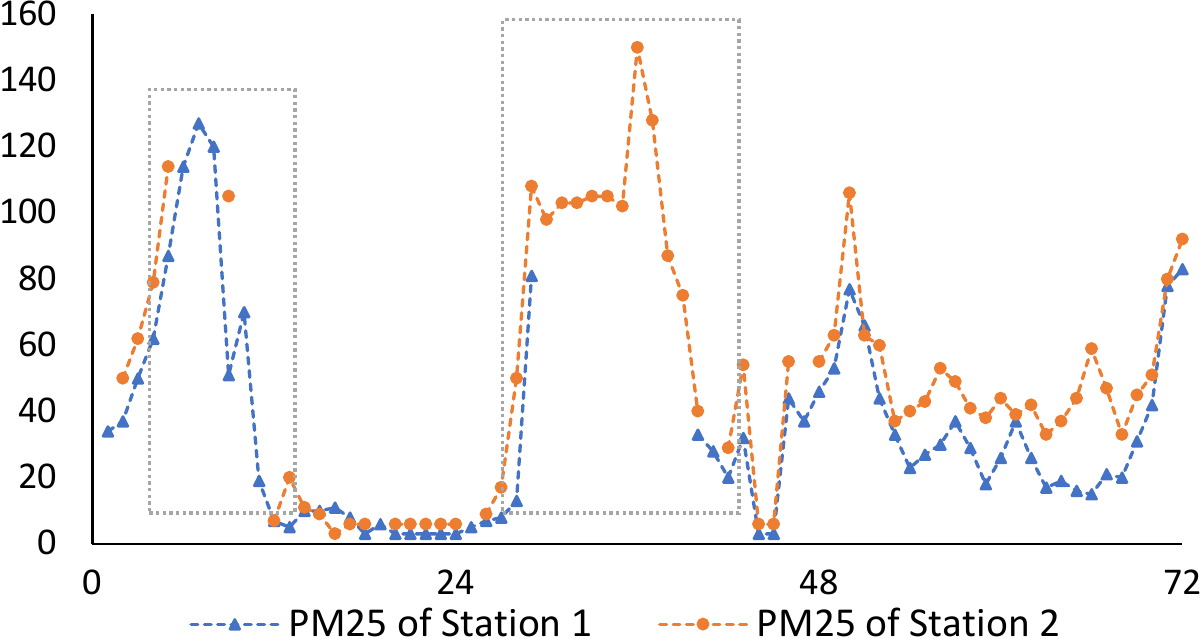}
    \caption{Two time series from Beijing air pollution dataset.}\label{fig:MTS}
    \label{missing_example} 
\end{center}
\end{figure}

In the real-world, MTS data are usually incomplete due to various reasons, such as broken sensors, failed data transmissions, or damaged storages. For example, Figure~\ref{fig:MTS} gives two multivariate time series snippets from Beijing air pollution data, both of which contain apparent missing values marked by gray boxes (i.e., 20 of the 144 data points are unobserved).
Missing values damage temporal dependencies in MTS sequences \cite{luo2018multivariate,cao2018brits}, make it hard to apply existing RNN-based models on incomplete sequences and increase the difficulty of MTS forecasting tasks. As shown in Figure~\ref{fig:MTS}, because of missing values, the second peak of the blue signal is not observed, and cannot be inferred by simply relying on existing RNNs.
Therefore, it is vital to design models that handle missing values in MTS data to perform accurate forecasting.
Many prior efforts have been dedicated to this direction. For example, two-step approaches that first omit or impute missing values then process time series forecasting based on the pre-processed data are explored in \cite{yi2016st,garcia2010pattern}. 
% A variety imputation methods can be applied, such as statistical features (e.g., mean, median, top frequency, or zero), matrix factorization \cite{yi2016st}, and EM algorithm \cite{garcia2010pattern}. 
% However, those two-step approaches fail to fully explore missing patterns for forecasting and can hurt the forecasting performance \cite{wells2013strategies}.
% To solve this issue, recent studies 
End-to-end solutions, where the missing patterns are modeled jointly with forecasting tasks, are investigated in \cite{che2018recurrent,cao2018brits,luo2018multivariate}. However, those methods only explore local statistical features, while the global temporal patterns in exogenous sequences are neglected.

Jointly modeling \textit{local} and \textit{global} temporal dynamics is very promising for MTS forecasting with missing values. 
Though constructing local statistics (e.g., empirical mean and last observations) to estimate missing variables have certain potential~\cite{che2018recurrent},
these local statistics are unreliable when the missing ratio raises up or  consecutive missing values occur as illustrated in Figure \ref{fig:MTS}.
This problem can be alleviated by adopting global temporal dynamics.
From a global perspective, there exist many MTS snippets with close temporal patterns. For example, in Figure~\ref{fig:MTS}, two MTS sequences from different air quality stations share similar temporal patterns. Although it is hard to recover consecutive missing values (i.e., circled by the grey dash boxes) purely from local statistics of one MTS, aggregating temporal patterns in both sequences is rather promising. The temporal patterns of one time series can be utilized for the other when dealing with missing values. 
However, how to take advantage of global temporal dynamics is a very challenging problem, which is under-explored in existing work.

To address the aforementioned challenges, we propose a novel framework \ours to model \textbf{L}ocal and \textbf{G}lobal temporal dynamics jointly for MTS forecasting. 
\ours absorbs model designs from previous work \cite{che2018recurrent}, where LSTM is leveraged for MTS forecasting.
Since the original LSTM is unable to handle incomplete input, we first construct estimations for missing values.
Specifically, representative local statistic features are constructed for each variable in an MTS. Besides, a memory module is designed for \ours to explicitly leverage knowledge from exogenous sequences to generate global estimations for missing values. This is achieved by using local statistics as keys to query a global optimized memory component. 
We further introduce adversarial training to enhance the modeling of global temporal distribution. A discriminator is built to identify the generated MTS from real samples. Meanwhile, \ours aims at producing forecasting sequences that are hard to be identified, which are also closer to the actual global distribution of real MTS data. The main contributions of the paper are: 

\begin{itemize}[leftmargin=*]
    \item We study a new problem of MTS forecasting with missing values by exploring local and global temporal dynamics.
    \item We propose a novel framework \ours, with a memory module to capture global temporal dynamics for missing values and adversarial training to enhances the modeling of global temporal distribution.
    \item We conduct extensive experiments on four large-scale real-world datasets to validate the proposed approach.
\end{itemize}

\section{Related Work} \label{sec:related}
% In this section, we briefly review related work, which includes MTS forecasting and MTS forecasting with missing values.

% \textbf{MTS Forecasting}
% {\color{blue} start with importance of MTS forecasting. then works MTS forecasting followed by works on MTS forecasting with missing values. Finally discuss the difference of our framework with existing work. Usually should be around half page}
Various methods have been proposed for MTS forecasting, such as Autoregressive (AR), Vector Autoregression (VAR), Autoregressive moving average (ARMA), standard regression models (e.g., support vector regression \cite{smola2004tutorial}, linear regression, and regression tree methods \cite{chen2016xgboost}).
Inspired by the recent success of deep neural networks, many RNN-based methods \cite{lai2018modeling,qin2017dual} are developed for MTS forecasting. Even some vanilla RNNs, such as GRU \cite{chung2014empirical} and LSTM \cite{hochreiter1997long}, can outperform the non deep learning models significantly \cite{chang2018memory}. However, \textit{none of those approaches can handle input with missing values}.

% \subsection{MTS Forecasting with Missing Values}
To handle missing values in MTS, the simplest solution would be removing all samples with missing values, such as pairwise deletion \cite{marsh1998pairwise}. Obviously, such methods \textit{ignore many useful information}, especially with a high missing ratio \cite{king1998list}.
General data imputation methods such as statistical imputation (e.g., mean, median), EM-based imputation \cite{nelwamondo2007missing}, K-nearest neighborhood \cite{friedman2001elements}, and matrix factorization \cite{friedman2001elements} can be applied for the unobserved variables. However, those general approaches \textit{fail to model temporal dynamics of time series}. 
Even if MTS imputation methods, such as multivariate imputation by chained equations \cite{azur2011multiple} and generative adversarial network \citeauthor{luo2018multivariate}, can be applied to fill in missing values first, \textit{training a forecasting model on pre-processed MTS data would lead to sub-optimal results, since the temporal patterns of missing values are totally isolated from forecasting models} \cite{wells2013strategies}.
To tackle this issue, some researchers propose end-to-end frameworks that jointly estimate missing values and forecast future MTS.
\citeauthor{che2018recurrent} introduce GRU-D that imputes missing values using the linear combination of statistical features.
\citeauthor{Yoon2017MultidirectionalRN} propose M-RNN that leverages bi-directional RNN for the imputation. 
\citeauthor{cao2018brits} model the relationships between missing variables to simultaneously perform imputation and classification/regression in one neural graph. 
However, \textit{those solutions focus on localized temporal dependencies and fail to model global temporal dynamics}.

% Our framework is inherently different from existing work: (i) We propose to study MTS forecasting with missing values from local and global temporal perspectives; and (ii) We design a new model with leverages LSTM, memory network and adversarial learning to simultaneously capture local and global temporal dynamics for MTS forecasting with missing values.

\section{Problem Formulation} \label{sec:problem}
In practice, many multivariate time series signals are sampled evenly. Thus, we assume time span is divided into equal-length time intervals. Let $\mathbf{X} = \{\mathbf{x}_1, \mathbf{x}_2, \dots, \mathbf{x}_n\} $ denote one MTS of length $n$, where $\mathbf{x}_i \in \mathbb{R}^{d}$ is the observation at the $i$-th time interval, $x_i^j$ is the $j$-th variable of $\mathbf{x}_i$, and $d$ is the number of variables.
Let mask matrix $\mathbf{M} = \{\mathbf{m}_1, \mathbf{m}_2, \dots, \mathbf{m}_n\}, \mathbf{m}_i \in \{0,1\}^{d}$ denote the missing status of each variable, where $m_i^j = 0$ if $x_i^j$ is unobserved/missing, otherwise, $m_i^j = 1$. Note that we can use a symbol to denote missing values in $\mathbf{X}$ (e.g., $null$). 

We are interested in a general MTS forecasting task. Given $N$ incomplete MTS observation $\{\mathbf{X}^j\}_{j=1}^{N}$ and their masks $\{\mathbf{M}^j\}_{j=1}^{N}$, we aim at learning a function $f$ that can forecast the values in future $k$ time intervals ($\{\mathbf{x}_{n+1}, \dots, \mathbf{x}_{n+k}\}$) of any new MTS, given its historical $n$ observations $\mathbf{X}$ with the mask matrix $\mathbf{M}$.

\section{The Proposed Framework}
\begin{figure}[t]
\begin{center}
    \includegraphics[width=.95\columnwidth]{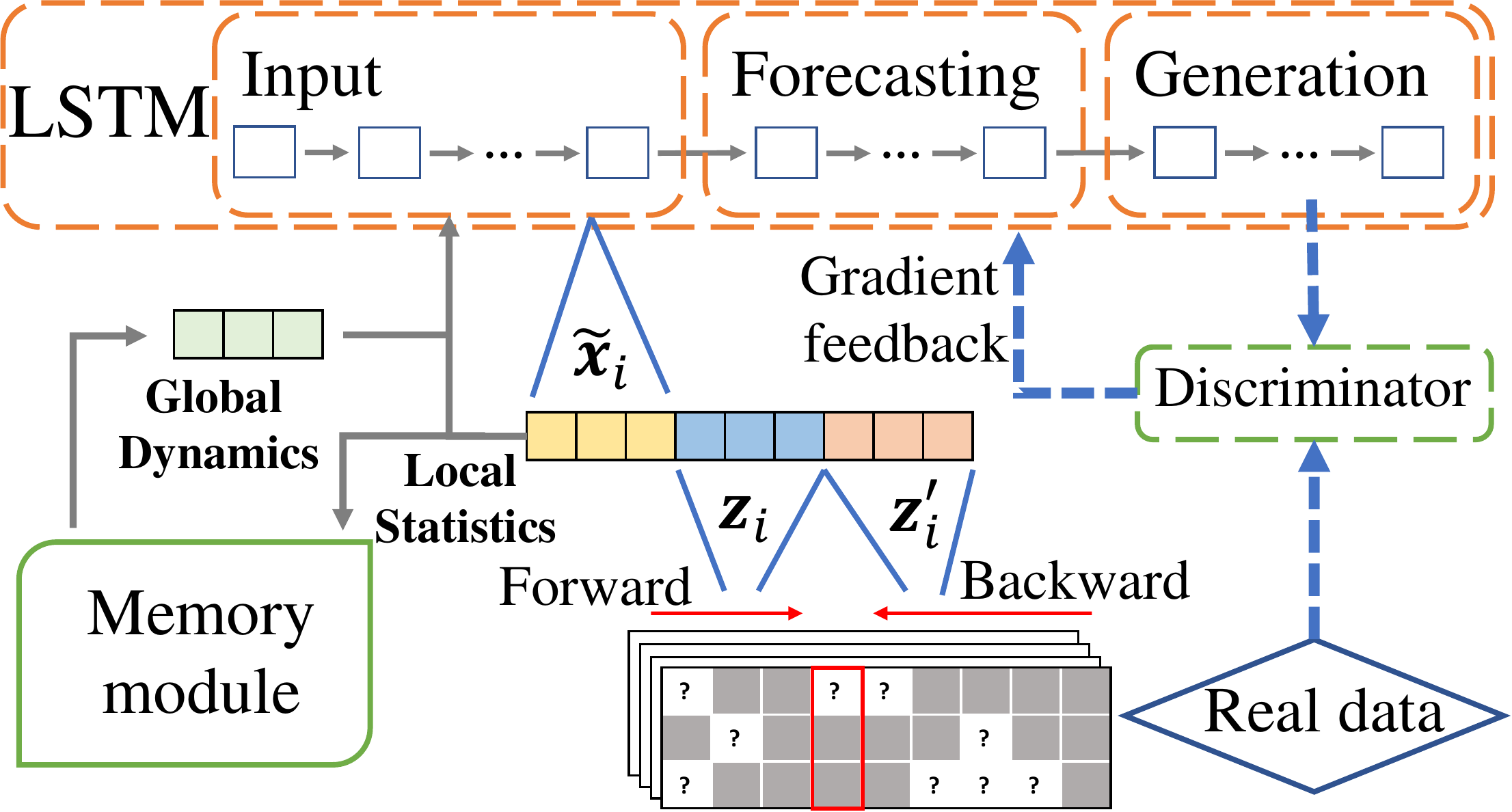}
    \caption{An illustration of \ours.
    }
    \label{framework} 
\end{center}
\end{figure}

Figure \ref{framework} illustrates the proposed framework \ours. \ours is built on LSTM to forecast future MTS values. 
We design a memory module which  contains temporal information from exogenous sequences to impute MTS during the running time of LSTM.
Specifically, we first extract local statistic features for every time interval, then use them as keys to query a memory component, which is jointly optimized with LSTM on all MTS data.
The query results, which preserve global temporal dynamics, are further combined with local statistic features to serve as the input of the LSTM.
% Despite the combined features solve the problem of missing values, their data distribution is different from real MTS data.
We also introduce adversarial training on forecasted sequences to make sure they follow the global distribution.
The whole framework of \ours is trained in an end-to-end manner.
Next, we introduce each module of \ours in detail.

\label{sec:framework}

\subsection{MTS forecasting with LSTM}
Recurrent neural networks (RNNs) have demonstrate remarkable success in various MTS forecasting tasks \cite{lai2018modeling,chang2018memory}. To leverage the advances of RNN, we build \ours on the top of Long short-term memory (LSTM) network, a variant of RNN which is able to capture long/short term dependency. 
Note that other RNN variants such as GRU \cite{cho2014learning} can serve as the replacement of the LSTM. 
Formally, LSTM takes one data point of the time series as input in each step, and iterates from $\mathbf{x}_1$ to $\mathbf{x}_n$. Suppose $\mathbf{x}_t$ is currently fed to the LSTM, $\mathbf{h}_{t-1}$ and $\mathbf{c}_{t-1}$ are the hidden state and memory cell of LSTM at previous step $t-1$, then the hidden state of memory cell at time $t$ can be calculated as:
\begin{equation*}
\small
    \begin{aligned}
        \mathbf{i}_{t} & =\sigma\big(\mathbf{x}_{t}\mathbf{W}^{i}+\mathbf{h}_{t-1}\mathbf{U}^{i}+\mathbf{b}^i\big),\ 
        \mathbf{f}_{t}  =\sigma\big(\mathbf{x}_{t}\mathbf{W}^{f}+\mathbf{h}_{t-1}\mathbf{U}^{f}+\mathbf{b}^f\big)\\
        \mathbf{o}_{t} & =\sigma\big(\mathbf{x}_{t}\mathbf{W}^{o}+\mathbf{h}_{t-1}\mathbf{U}^{o}+\mathbf{b}^o\big),\ 
        \tilde{\mathbf{c}}_{t}  =\tanh\big(\mathbf{x}_{t}\mathbf{W}^{c}+\mathbf{h}_{t-1}\mathbf{U}^{c}\big)\\
        \mathbf{c}_{t} & =f_{t}\odot \mathbf{c}_{t-1}+i_{t}\odot \tilde{\mathbf{c}}_{t},  \quad\quad\quad\ 
        \mathbf{h}_{t}  =\tanh(\mathbf{c}_{t})\odot \mathbf{o}_{t}
    \end{aligned}
\end{equation*}
where $\mathbf{i}_t$ is the input gate, $\mathbf{f}_t$ is the forget gate, $\mathbf{o}_t$ is the output gate, $\odot$ is the element-wise product, $\sigma$ is the sigmoid function, and $\mathbf{W}^{*}$, $\mathbf{U}^{*}, \mathbf{b}^{*}$ are parameters. Given the current hidden state $\mathbf{h}_t$, the forecasting of next data point $\tilde{\mathbf{x}}_{t+1}$ can be generated recurrently as:
\begin{equation} \label{lstm_output}
    \tilde{\mathbf{x}}_{t+1} = \mathbf{h}_{t}\mathbf{U}+\mathbf{b}.
\end{equation}
However, the original LSTM cannot handle missing values in its input. Obviously, if $\mathbf{x}_t$ contains unobserved variables, matrix productions such as $\mathbf{x}_{t}\mathbf{W}^{i}$ are invalid.
One solution is using $\tilde{\mathbf{x}}_{t}$ as an alternative of $\mathbf{x}_t$. However, early errors in $\tilde{\mathbf{x}}_{t}$ can be quickly amplified in the following steps \cite{bengio2015scheduled}, leading to inaccurate forecasting.

Therefore, appropriate estimations of missing values should be constructed for the LSTM. We tackle the problem by exploring temporal dynamics from both local and global perspectives with a memory module.
In the next section, we discuss its technical details.

\subsection{Memory Module}
\begin{figure}
\begin{center}
    \includegraphics[width=.70\columnwidth]{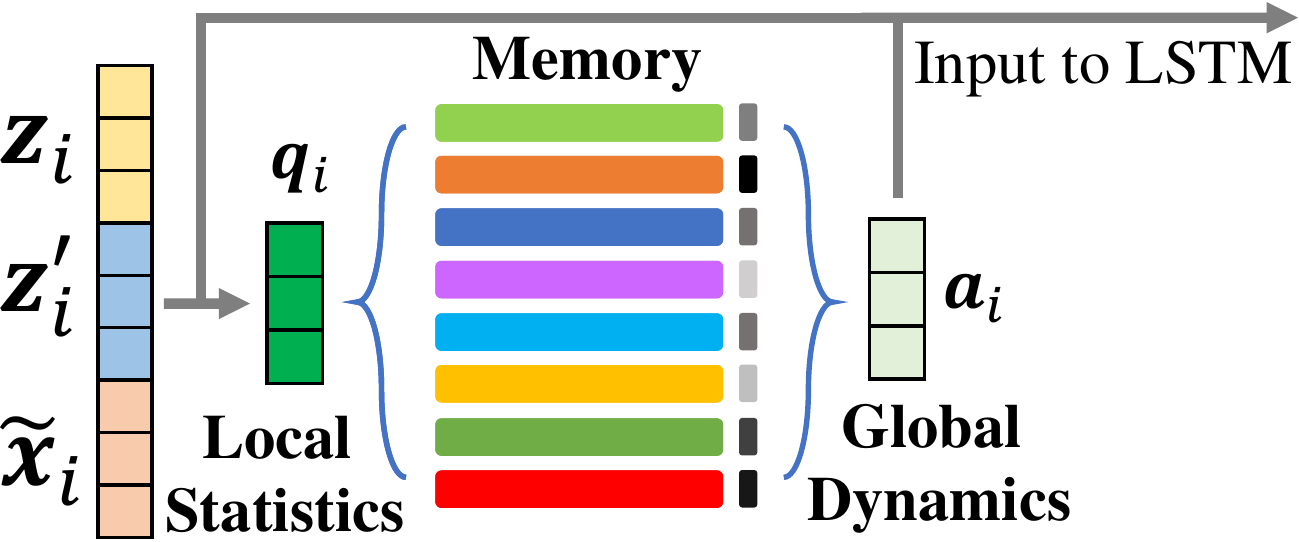}
    \caption{Architecture of the memory module.}
    \label{memory_fig}
\end{center}
\end{figure}

The basic idea of the memory module is to learn a parameterized memory which caches global temporal patterns and projects each variable to the same feature space with the memory. 
For each variable in a MTS, we first capture informative statistics from the local context of this time series, then leverage local statistics as keys to query the memory component, which returns representation vectors with global temporal dynamics.
The memory module brings two advantages: (i) learn and store meaningful temporal patterns from a global perspective; and (ii) utilize the knowledge of temporal patterns to construct global representations. \textit{Note that the memory module is not the memory cell of the LSTM as shown in Figure~\ref{framework}} and \ref{memory_fig}.

\subsubsection{Capturing Local Statistics}
We extract useful local statistic features using the contextual information from observed parts of the time series for missing values.
Following prior studies \cite{che2018recurrent}, we first generate empirical mean and last observation of every time stamp as follows:

\textit{Empirical Mean}: for variable $x_i^j$,
we construct its empirical mean using all available observations of $x_*^j$ before time $i$, i.e., $\bar{x}_i^j = \sum_{l=1}^{i-1} {m_l^j  x_l^j} \big/ \sum_{l=1}^{i-1} m_l^j$.
The mean of previous observations reflects the time-aware data distribution of $x_i^j$ and serves as the prior knowledge of the variable.

\textit{Last Observation}: the last observation of $x_i^j$ is the first available $j$-th variable before time interval $i$, which is the most temporally close neighbor. We use $\dot{x}_i^j$ to denote the last observation of $x_i^j$. Note that $\dot{x}_i^j$ isn't necessary equal to ${x}_{i-1}^j$ because ${x}_{i-1}^j$ could also be missing.
We further introduce an indicator $\mathbf{\delta} \in \mathbb{R}_{+}^{d \times n}$ to record the temporal distance between each $x_i^j$ and its last observation, which reflects the confidence and trustworthy of previous values:
\begin{equation}
\small
\delta_i^j=
\begin{cases}
  0, & \text{if}\ i = 1 \text{ or } m_i^j = 1\\
  \delta_{i-1}^j + 1, & \text{if}\ m_i^j = 0
\end{cases}.
\end{equation}
Generally, when $\delta_i^j$ is small, we tend to trust $\dot{x}_i^j$ more; and when $\delta_i^j$ becomes larger, the averaged value $\bar{x}_i^j$ would be more representative.
Based on the above assumption, we propose the following decaying mechanism to balance empirical mean and last observation:
\begin{equation}
    \mathbf{\gamma}({\delta}_i^j) = \exp{(-\max(0, {{w}^j} {\delta}_i^j + b^{j}))},
\label{delta}
\end{equation}
where $w^j$ and $b^j$ are parameters.
The above decay mechanism leverages an exponentiated negative rectifier so that the decay value $\gamma$ is monotonous decreasing in the range between 0 and 1 w.r.t ${\delta}_i^j$  \cite{che2018recurrent,luo2018multivariate}.
The localized estimation for $x_i^j$ is constructed as follows:
\begin{equation}
    z_i^j \leftarrow m_i^j x_i^j +  (1 -  m_i^j) [ \mathbf{\gamma}(\delta_i^j) \dot{x}_i^j + (1 -  \mathbf{\gamma}(\delta_i^j)) \bar{x}_i^j ].
\label{impu}
\end{equation}
We use $\mathbf{z}_i=[z_i^1, \dots, z_i^d]$ to denote local statistic features for $\mathbf{x}_i$. For observed variables, their original values are directly used. For missing values, we combine empirical mean with last observation to construct $\mathbf{z}_i$.

However, $\mathbf{z}_i$ only takes data points observed before the $i$-th time interval into consideration.
Similar local statistics can also be extracted from time interval $i+1$ to $n$. 
This is achieved by first reverse $\mathbf{X}$ and $\mathbf{M}$ on the temporal dimension, then extracting local statistics following the same definition of $\mathbf{z}_i$ on time interval $1$ to $n-i+1$. % (which equals to time interval $i+1$ to $n$ before reversing).
We use $\mathbf{z}_i^\prime$ to denote local statistics from time interval $i+1$ to $n$. 
As shown in Figure \ref{framework}, $\mathbf{z}_i$ and $\mathbf{z}_i^\prime$ are forward and backward local statistic features, respectively.

In addition to the forward and backward local statistic features, the LSTM naturally provides estimations for missing variables. %For example, when it recursively reads the values at each observation, the hidden state is updated using the previous memory of LSTM and new inputs, which captures local temporal pattern. 
Specifically, we follow Equation \ref{lstm_output} and use the output of LSTM at the previous time step $i-1$ as another local statistic from a model view:
\begin{equation} \label{eq:model_estimation}
    \tilde{\mathbf{x}}_{i} = \mathbf{h}_{i-1}\mathbf{U}+\mathbf{b}
\end{equation}

\subsubsection{Modeling Global Dynamics}
The above imputations $\mathbf{z}_i$, ${\mathbf{z}}_i^\prime$, and $\tilde{\mathbf{x}}_{i}$ can fed directly in LSTM to train a MTS forecasting model \cite{che2018recurrent}. However, such an approach is sub-optimal. $\mathbf{z}_i$ and ${\mathbf{z}^\prime}_i$ become less trustful as the missing ratio raises up. Besides, purely relying on local statistics ignores global temporal dynamics from exogenous sequences, which potentially benefit the estimation of missing values.
It is likely to capture time series snippets/patterns from other sequences that are temporally similar (e.g., periodicity) to the contextual of a missing value.
For example, to impute one missing data point in a trajectory, similar time series snippet may be found from other trajectories.

However, capturing such global temporal dynamics to find informative temporal patterns for missing values is very challenging. Simply comparing with all potential snippets to find similar ones is impractical due to high computational costs.
Recently, memory network \cite{weston2014memory} has shown promising results in capturing patterns for sequential data \cite{sukhbaatar2015end,chang2018memory}. Generally, a memory network initializes a memory component to store feature representations that optimized explicitly on the whole dataset. 
Those stored representations can be retrieved and utilized for specific tasks \cite{tang2017end,kumar2016ask}.
We design a memory module to  capture global temporal dynamics explicitly, as shown in Figure \ref{memory_fig}.
We assume there are $L$ temporal patterns existing in the dataset ($L$ is a hyperparameter), and initialize a parameterized memory $\mathcal{G} \in \mathbb{R}^{L \times d^\mathcal{G}}$, where $d^\mathcal{G}$ is the dimension of pattern representation. The memory $\mathcal{G}$ is updated jointly with the LSTM.
We utilize local statistics as keys to query the memory module because they can represent the uniqueness of variables. Specifically, queries to the memory module are constructed as follows:
\begin{equation}
\label{query}
\small
    \mathbf{q}_i = \mathbf{W}_q [\mathbf{z}_i || \mathbf{z}_i^\prime || \tilde{\mathbf{x}}_i] + \mathbf{b}_q,
\end{equation}
where $||$ denotes concatenation on column. $\mathbf{W}_q$ and $\mathbf{B}_q$ are parameters. Then we calculate the similarity between $\mathbf{q}_i$ and the memory component $\mathcal{G}$:
\begin{equation}
\small
    \mathbf{s}_i = \text{Softmax}(\mathcal{G} \cdot \mathbf{q}_i).
\end{equation}
The similarity scores measure the importance of each temporal patterns in the memory. Any pattern with a higher attention score is more similar to the context of targeting missing value. The representation vector of $\mathbf{x}_i$ that preserves global temporal dynamics is then constructed from the weighted sum of all temporal patterns in $\mathcal{G}$:
\begin{equation} \label{memory}
\small
    \mathbf{a}_i = \sum_{l=1}^{L} s_i^l \mathcal{G}(l),
\end{equation}
\noindent{}where $\mathcal{G}(l)$ represents the $l$-th row of the memory component. Besides, since variables at the same time interval interact with each other in Equation \ref{query}, $\mathbf{a}_i$ also preserves inner correlations of variables at the same time interval.
We combine local statistic features and global representations to construct the input of LSTM. Specifically, $\mathbf{z}_i$, ${\mathbf{z}}_i^\prime$, $\tilde{\mathbf{x}}_{i}$, and $\mathbf{a}_i$ are averaged as the input at time interval $i$.
Note that some of the local statistics can become unavailable in some cases. For example, we cannot construct $\mathbf{z}_i$ for the first missing value, and no $\mathbf{z}_i^\prime$ is available for the forecasting stage. We set unavailable local statistics to $\mathbf{0}$. However, we can generate either the forward or the backward local statistic features unless the whole time series is empty. This ensures \ours is more reliably than those purely using the forward local statistic \cite{che2018recurrent}.
The forecasting results ${\mathbf{X}}_p = [ \tilde{\mathbf{x}}_{n+1}, \dots, \tilde{\mathbf{x}}_{n+k}]$ are generated after $n$-th iteration of LSTM.
The aligned ground truth data for ${\mathbf{X}}_p$ is denoted as $\hat{\mathbf{X}}_p$, which also contains missing values. 
Therefore, we incorporate the mask matrix into mean-square-error and propose the following objective function to train \ours:
\begin{equation} \label{loss1}
\small
    \min_{\theta} \mathcal{L}_p(\theta) = \frac{1}{N} \sum_{j=1}^N  \|(\mathbf{X}_p^j - \hat{\mathbf{X}}_p^j) \odot \mathbf{M}_p^j \|_2,
\end{equation}
\noindent{}where $\theta$ are parameters of \ours, including parameters of the LSTM and the memory component, $\mathbf{M}_p^j$ is the mask matrix of the $j$-th MTS data sample $\mathbf{X}^j$ over the predicted variables, and $\odot$ is dot-production. Because of the mask matrix, \ours is optimized over the observed part of $\hat{\mathbf{X}}_p^j$.
% {\color{blue} check if this makes sense, and revise accrodingly to make it clear that we have N MTS data samples}

\subsection{Adversarial Training}
% Recently, generative adversarial network (GAN) shown its ability in estimating the underlying data distribution, which paves a way to \cite{goodfellow2014generative} has been successfully applied to image generation \cite{radford2015unsupervised} and video prediction \cite{liang2017dual}. Inspired by its idea, we design an extra unsupervised loss function. 
The objective function of \ours in Equation \ref{loss1} only considers available variables. When the missing ratio is relatively high, the proposed objective function becomes inefficient because most values of $\mathbf{M}_p$ is zero when sampling under the same data distribution. %Moreover, training the whole framework on few objective could bring bias to the computed loss value, thus hurt the model training. %To tackle the above limitation, we propose an adversarial training schema that incorporate extra guidance to the framework.
The predicted future sequences should also follow the same data distribution of the true MTS data.
If we can encourage \ours to generate more realistic data distribution, the overall accuracy of MTS forecasting can be improved.
To achieve this goal, we introduce adversarial training to control the distribution of generated MTS.

Recently, generative adversarial networks (GANs) \cite{goodfellow2014generative} have been widely applied to various domains  \cite{yu2017seqgan,sun2019megan,shu2018deep}. Typical GAN consists of a generator and a discriminator.
The discriminator tries to distinguish samples from the generator and those from read data.
The generator tries to generate samples that can ``fool'' the discriminator by modeling  data distribution %(i.e., making it hard to tell whether the sample is ``fake'' or ``true'').
With such a min-max game, the generator can create more realistic samples. % following distribution $P_g$ which is similar to real data distribution $P_\text{data}$.% i.e.,  $P_g \approx P_\text{data}$.
% By training generator and discriminator in turn, the generator receives gradient feedback from the discriminator and learns better data distribution. 
This motivates us to adopt adversarial learning to enhance the forecasting. %for incorporating the global data distribution so that more realistic MTS are generated as the forecasting.
We design a discriminator $D$, as illustrated in  Figure~\ref{framework}. LSTM generates future sequences as the forecasting %conditioned on previous observations 
to ``fool'' the discriminator $D$; while $D$ is trained to identify whether the input sequence %(i.e., the forecasting from the LSTM) 
is fake. Through iterative training, the LSTM is more capable of generating time series that fit the underlying distribution~\cite{goodfellow2014generative}, which makes the forecasting result more accurate. %Note that different from discrete sequences such as text and check-ins, multivariate time series are continuous real values, and the generator (i.e., LSTM) can receive gradient from the discriminator \cite{fedus2018maskgan}.

Specifically, we adopt W-GAN \cite{arjovsky2017wasserstein} and construct a two-layer convolution net as $D$. Given a MTS $s$ as input, $D$ outputs a real value $D(s)$, which is higher if $s$ is real, and lower if $s$ is ``fake''. %Namely, $D$ aim at identifying the discrepancy between real and fake sequence.
The ``fake'' multivariate time series of length $k^\prime$ are generated after the forecasting part. 
Let $\mathcal{X} = \{[\tilde{\mathbf{x}}_{n+k+1}, \dots, \tilde{\mathbf{x}}_{n+k+k^\prime}]\}$ denote a generated (fake) time series. 
To compile a ``true'' dataset that preserves latent data distribution, we sample a subset of complete time series snippets with same length $k^\prime$ from the raw dataset. Let $\mathcal{S}$ denote the sampled subset of time series snippets. Empirically, it is not a difficult task when $k^\prime$ is small (i.e., $k^\prime \le 5$).
The training objective of the discriminator is:
\begin{equation}
    \min_{\theta_D}\mathcal{L}_D = \min_{\theta_D} - \mathop{\mathbb{E}}_{s \sim \mathcal{S}} D(s) + \mathop{\mathbb{E}}_{s \sim \mathcal{X}} D(s), % {\color{blue} \text{add min/max here}}
\end{equation}
where $\sim$ denotes ``sampling from'', and $\theta_D$ is parameters of the discriminator. Generally, time series with a high probability of being a ``true'' sample will receive a higher score.
To generate more realistic sequences, the objective function for the LSTM is defined as:
\begin{equation} \label{train_d}
    \min_{\theta} \mathcal{L}_a = \min_{\theta} -\mathop{\mathbb{E}}_{s \sim \mathcal{X}} D(s),
\end{equation}
which aims at faking the discriminator.
Note that there is no overlap between the forecasting and the generated part, as we imperially find that adding adversarial loss on the forecasting part may hurt the performance. A potential reason is that the best time series to ``fool'' the discriminator might not be the most accurate forecasting result. Therefore, we put $\mathcal{L}_a$ on extra generated sequences to achieve the best performance. \citeauthor{luo2018multivariate} state a similar conclusion.

\subsection{Objective Function and Training}
We define the overall objective function to learn model parameters $\theta$ for an accurate MTS forecasting with adversarial training as follows:
\begin{equation*} \label{train_g}
% \small
\begin{aligned}
    \theta, \theta_D = \min_{\theta}\mathcal{L}_p +  \lambda \big [\max_{\theta_D} \mathop{\mathbb{E}}_{s \sim \mathcal{S}} D(s) -\mathop{\mathbb{E}}_{s \sim \mathcal{X}} D(s) \big],
\end{aligned}
\end{equation*}
where $\lambda$ balances the MTS forecasting part and the adversarial training part.

We use stochastic gradient descent to update model parameters.
The discriminator and the LSTM are trained alternatively until converged. We first update $\theta_{D}$ with real MTS snippets and generated ones, then optimize $\theta$ for the LSTM and the memory module while fixing $\theta_{D}$.
% The detailed training procedure of \ours is given in Algorithm 1 in the \textit{supplementary material}. 

\section{Experiment}\label{sec:experiment}
In this section, we present experiments to evaluate the proposed framework \ours. Specifically, we aim at answering the following research questions: (i) \textbf{RQ1}: Can \ours improve the accuracy of MTS forecasting with missing values? (ii) \textbf{RQ2} How robust is \ours w.r.t different missing ratios? (iii) \textbf{RQ3} How the memory module benefits \ours? (iv) \textbf{RQ4} How adversarial training contributes to \ours?
Next, we start by introducing various experiments on MTS forecasting to answer the above questions.

\subsection{Datasets}
Four large-scale real-world MTS datasets from different domains are selected to validate \ours:
\textbf{Beijing Air\footnote{https://www.kdd.org/kdd2018/kdd-cup}}: This dataset is introduced by KDD Cup 2018. We extract PM2.5 values from 35 monitoring stations in Beijing, and formulate multivariate time series. The values are reported  every hour between 05/01/2014 and 04/30/2015. It has a missing rate of 13\% over the temporal dimension. We use past 9-hour observations to train each model, and forecast PM2.5 values for the following 3 hours.
\textbf{PhysioNet}: PhysioNet \cite{silva2012predicting} provides 4000 multivariate clinical time series from intensive care unit (ICU). Each time series records 35 measurements such as glucose and heart-rate from the first 48 hours since the patient entered the hospital.
We compile time series from 12 important measurements such as heart-rate and temperature.
The missing ratio of PhysioNet is about 78\% over the temporal dimension.
We use the past 6 observations to forecast values in the coming 3 hours.
\textbf{Porto Taxi\footnote{https://www.kaggle.com/c/pkdd-15-predict-taxi-service-trajectory-i}}: This dataset includes approximately one million trajectories for 442 taxis running in the city of Porto during a complete year (from 01/07/2013 to 30/06/2014). Each trajectory contains many GPS coordinates (i.e., longitude and latitude) recorded chronologically. The sampling speed is 15 second per coordinates. We use the past 7 GPS coordinates to forecast the location of future points.
\textbf{London Weather\footnotemark[1]}: The dataset includes temperature, pressure, humidity, wind direction, and wind speed from 861 regions in London from 01/01/2017 to 03/27/2018. All features are collected hourly. We use the past 5 observations to forecast the coming values.

The first and second datasets naturally contain missing values, which are used as real-world settings to answer the first question. For the rest two datasets, we randomly remove $p$\% of observed values ($p \in \{10, 30, 50, 70, 90\}$) to study the robustness of \ours and compared methods.

\subsection{Compared Baselines}
We compare \ours with classical and state-of-the-art baselines, including two non-RNN methods, two time series imputation methods, and two RNN methods:
\begin{itemize}[leftmargin=*]
    \item \textbf{Linear Regression} (LR): Because conventional linear regression model cannot directly handle missing values, we concatenate each MTS with its mask matrix as the input features to train LR for the forecasting task.
    \item \textbf{XGBoost} \cite{chen2016xgboost}: XGBoost is widely used in time series analysis and machine learning fields. We use the same setting of LR to train XGBoost.

    \item \textbf{MICE}~\cite{azur2011multiple}: MICE fills the missing values using multiple imputations with chained equations. We first apply MICE to impute miss values. We then train LSTM for the forecasting task.
    
    \item \textbf{GRUI} \cite{luo2018multivariate}: GRUI leverages GAN and RNN for time series imputation. Similar to MICE, we first train GRUI for time series imputation. Then we train LSTM on imputed data as the forecasting model.
    
    \item \textbf{GRU-D} \cite{che2018recurrent}: GRU-D combines statistical features and linear decay in RNN to tackle missing variables in time series. It is proposed for multivariate time series forecasting task.
    
    \item \textbf{BRITS} \cite{cao2018brits}: BRITS designs bi-direction recurrent neural architecture for time series imputation and forecasting. It models missing patterns explicitly and improves forecasting accuracy.
    %{\color{blue} how do you modify?}
\end{itemize}

\subsection{Experimental Settings}
We normalize each dataset and ensure all time series variables have the same scale (i.e.,  mean and variance) on each dataset so that their averaged results are comparable.
For each dataset, we randomly select 70\% of MTS data for training, 10\% for validation to tune hyperparameters, and the remaining 20\% for testing. 
We set the dimension of the hidden unit to 32 for the LSTM. We select $L$ from $\{8, 16, 32, 64\}$ to create the memory component according to the performance on validation sets. 
The dimension of each memory slot is 128. We tune $\lambda$ that balance MTS forecasting and adversarial training on validation sets for the best performance.
The discriminator contains two convolutional layers following by two fully-connected layers. $3 \times 3$ kernels are used for both convolutional layers. The channel sizes are 64 and 128 for the first and second convolutional layer, respectively. The dimensions of fully connected layers are 1024 and 1.

Two widely used evaluation metrics, i.e.,  \textit{root mean square error} (RMSE) and \textit{mean absolute error} (MAE), are adopted. 
%The detailed definitions are given as:
%$$RMSE=\sqrt{\frac{1}{N}\sum_{i}(\hat{y}_i-y_i)^2},\ \ MAE=\frac{1}{N}\sum_{i}{|\hat{y}_i-y_i|},$$
%where $\hat{y}_i$ is the ground-truth of a future value, $y_i$ is the forecasting by the model, and $N$ denotes total number of values. 
Since different variables have different scales, we report the RMSE and MAE on their normalized values. \textit{The smaller RMSE and MAE are, the better the performance is.}

\subsection{Performance Comparisons}

\begin{table*}[!h]
    \centering
\footnotesize
\caption{MTS forecasting performances on Beijing Air and PhysioNet.}
    \label{tab:overall}
    \resizebox{1.95\columnwidth}{!}{
    \begin{tabular}{c|c|cccccc}
\hline \hline
     \multirow{2}{*}{Datasets}   & k                    & \multicolumn{2}{c}{1}                 & \multicolumn{2}{c}{2}                 & \multicolumn{2}{c}{3}                 \\ \cline{2-8}
                    & Metric               & RMSE              & MAE               & RMSE              & MAE               & RMSE              & MAE               \\ \hline 
\multirow{7}{*}{Beijing Air} & LR                   & 0.0398$\pm$0.0018 & 0.0261$\pm$0.0003 & 0.0550$\pm$0.0014 & 0.0371$\pm$0.0001 & 0.0661$\pm$0.0015 & 0.0454$\pm$0.0007 \\
                             & XGB                  & \textbf{0.0389$\pm$0.0004} & \textbf{0.0229$\pm$0.0017} & 0.0542$\pm$0.0010 & 0.0376$\pm$0.0015 & 0.0663$\pm$0.0009 & 0.0406$\pm$0.0013 \\
                             & MICE                 & 0.0646$\pm$0.0001 & 0.0417$\pm$0.0004 & 0.0703$\pm$0.0012 & 0.0452$\pm$0.0017 & 0.0766$\pm$0.0018 & 0.0493$\pm$0.0008 \\
                             & GRUI                 & 0.0601$\pm$0.0012 & 0.0387$\pm$0.0009 & 0.0667$\pm$0.0001 & 0.0433$\pm$0.0001 & 0.0754$\pm$0.0007 & 0.0502$\pm$0.0012 \\
                             & GRU-D                & 0.0459$\pm$0.0014 & 0.0305$\pm$0.0001 & 0.0554$\pm$0.0020 & 0.0366$\pm$0.0011 & 0.0649$\pm$0.0003 & 0.0432$\pm$0.0001 \\
                             & BRITS                & 0.0501$\pm$0.0001 & 0.0335$\pm$0.0012 & 0.0570$\pm$0.0003 & 0.0384$\pm$0.0008 & 0.0707$\pm$0.0017 & 0.0493$\pm$0.0002 \\
                             & \ours & 0.0451$\pm$0.0010 & 0.0300$\pm$0.0009 & \textbf{0.0519$\pm$0.0008} & \textbf{0.0332$\pm$0.0006} & \textbf{0.0597$\pm$0.0002} & \textbf{0.0373$\pm$0.0006} \\ \hline
\multirow{7}{*}{PhysioNet}   & LR                   & 0.2401$\pm$0.0003 & 0.1839$\pm$0.0006 & 0.2520$\pm$0.0009 & 0.1948$\pm$0.0007 & 0.2622$\pm$0.0001 & 0.1997$\pm$0.0004 \\
                             & XGB                  & 0.2308$\pm$0.0004 & 0.1753$\pm$0.0006 & 0.2481$\pm$0.0005 & 0.1913$\pm$0.0019 & 0.2598$\pm$0.0017 & 0.1972$\pm$0.0001 \\
                             & MICE                 & 0.1113$\pm$0.0000 & 0.0783$\pm$0.0011 & 0.1148$\pm$0.0008 & 0.0793$\pm$0.0006 & 0.1116$\pm$0.0018 & 0.0789$\pm$0.0019 \\
                             & GRUI                 & 0.1142$\pm$0.0017 & 0.0776$\pm$0.0012 & 0.1176$\pm$0.0001 & 0.0812$\pm$0.0008 & 0.1270$\pm$0.0005 & 0.0813$\pm$0.0016 \\
                             & GRU-D                & 0.1125$\pm$0.0013 & 0.0998$\pm$0.0003 & 0.1202$\pm$0.0011 & 0.0796$\pm$0.0014 & 0.1348$\pm$0.0002 & 0.0971$\pm$0.0011 \\
                             & BRITS                & 0.1082$\pm$0.0002 & 0.0720$\pm$0.0010 & 0.1158$\pm$0.0012 & 0.0734$\pm$0.0018 & 0.1158$\pm$0.0001 & 0.0785$\pm$0.0003 \\
                             & \ours & \textbf{0.1021$\pm$0.0004} & \textbf{0.0706$\pm$0.0002} & \textbf{0.0998$\pm$0.0002} & \textbf{0.0713$\pm$0.0002} & \textbf{0.1080$\pm$0.0005} & \textbf{0.0762$\pm$0.0007} \\
                             \hline
\end{tabular}}
\end{table*}

To answer \textbf{RQ1}, we compare \ours with baselines on Beijing Air and PhysioNet, where missing values naturally exist. We report the performance on the two datasets for $k=1,2,3$ (forecasting horizon) in  Table~\ref{tab:overall}, and make the following observations: \textbf{\textit{(i)}} \ours outperforms all the baseline methods for the majority of the cases, which shows the effectiveness of the memory module and adversarial learning for multivariate time series forecasting with missing values. The memory module explores global temporal dynamics and generates appropriate estimations for missing values; \textbf{\textit{(ii)}} when $k$ increases, i.e., when forecasting far future values, the performance of all the methods decreases, which is reasonable because it's more difficult to forecast far future values than near ones. However, \ours still significantly outperform the compared methods, which is because we adopt adversarial training on the predicted sequences to make the forecasting more realistic; \textbf{\textit{(iii)}} In addition, the performance improvement of \ours is much more significant on PhysioNet than Beijing Air. Compared with Beijing Air, PhysioNet has a higher missing ratio, which challenges the baseline methods; while \ours can still handle such high missing ratio, which further implies the effectiveness of \ours by designing memory network and adopting adversarial training.

\subsection{Robustness of \ours}
\begin{figure}[t]
        \centering
        \begin{subfigure}[b]{.23\textwidth}
            \centering
            \includegraphics[width=\columnwidth]{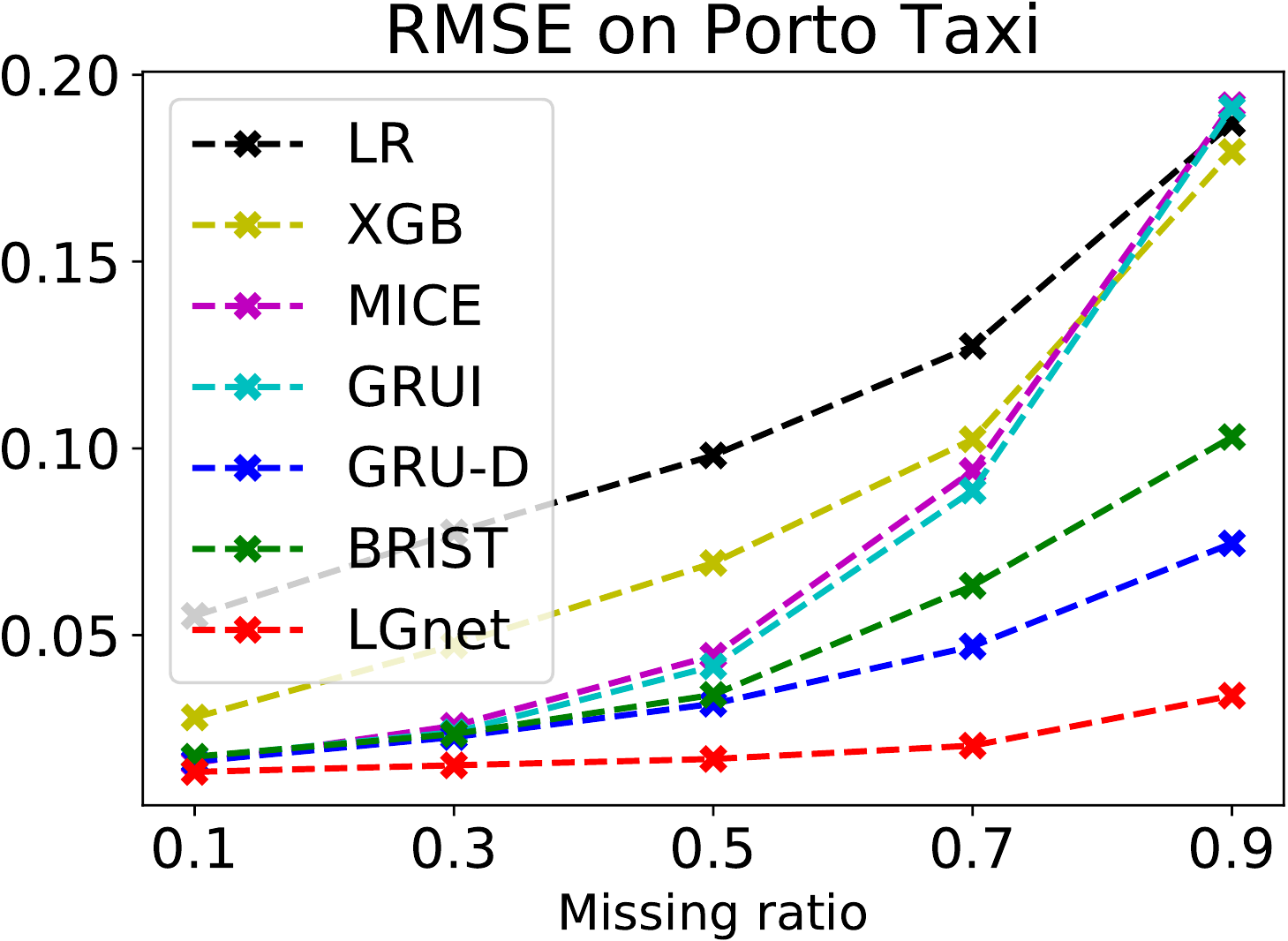}
        \end{subfigure}
        \begin{subfigure}[b]{.23\textwidth}  
            \centering 
            \includegraphics[width=\columnwidth]{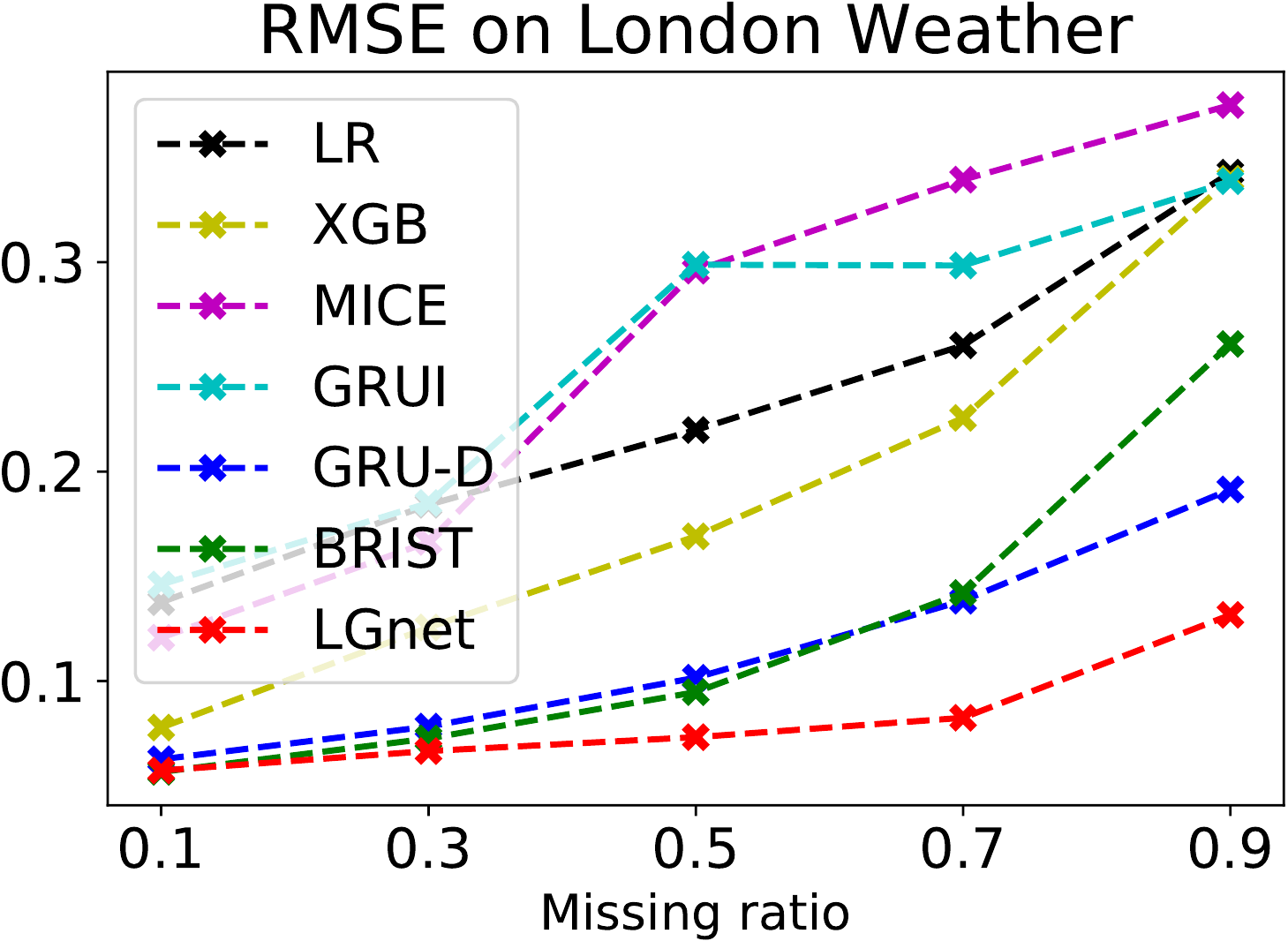}
        \end{subfigure}
        \begin{subfigure}[b]{.23\textwidth}   
            \centering 
            \includegraphics[width=\columnwidth]{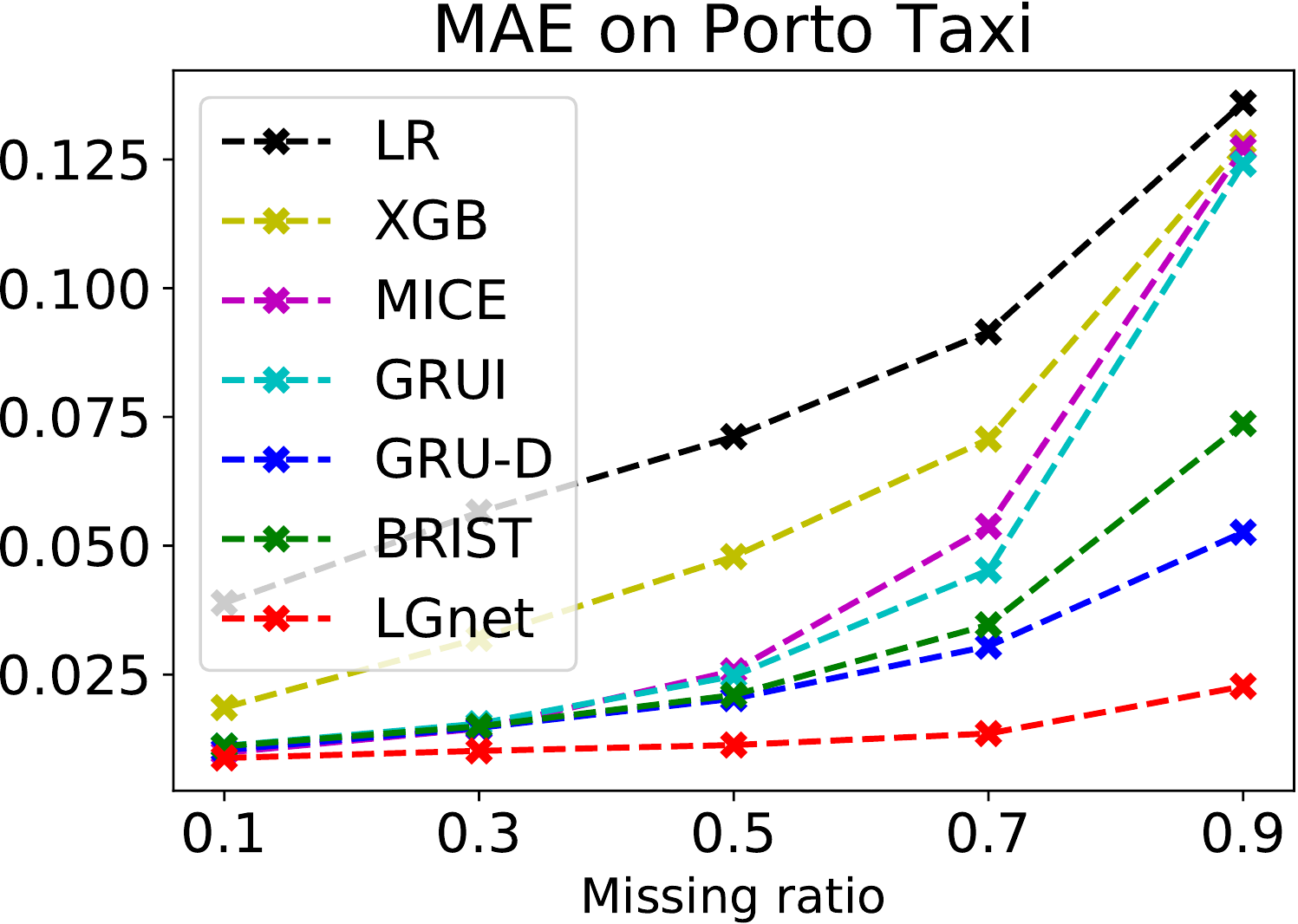}
        \end{subfigure}
        \begin{subfigure}[b]{.23\textwidth}
            \centering 
            \includegraphics[width=\columnwidth]{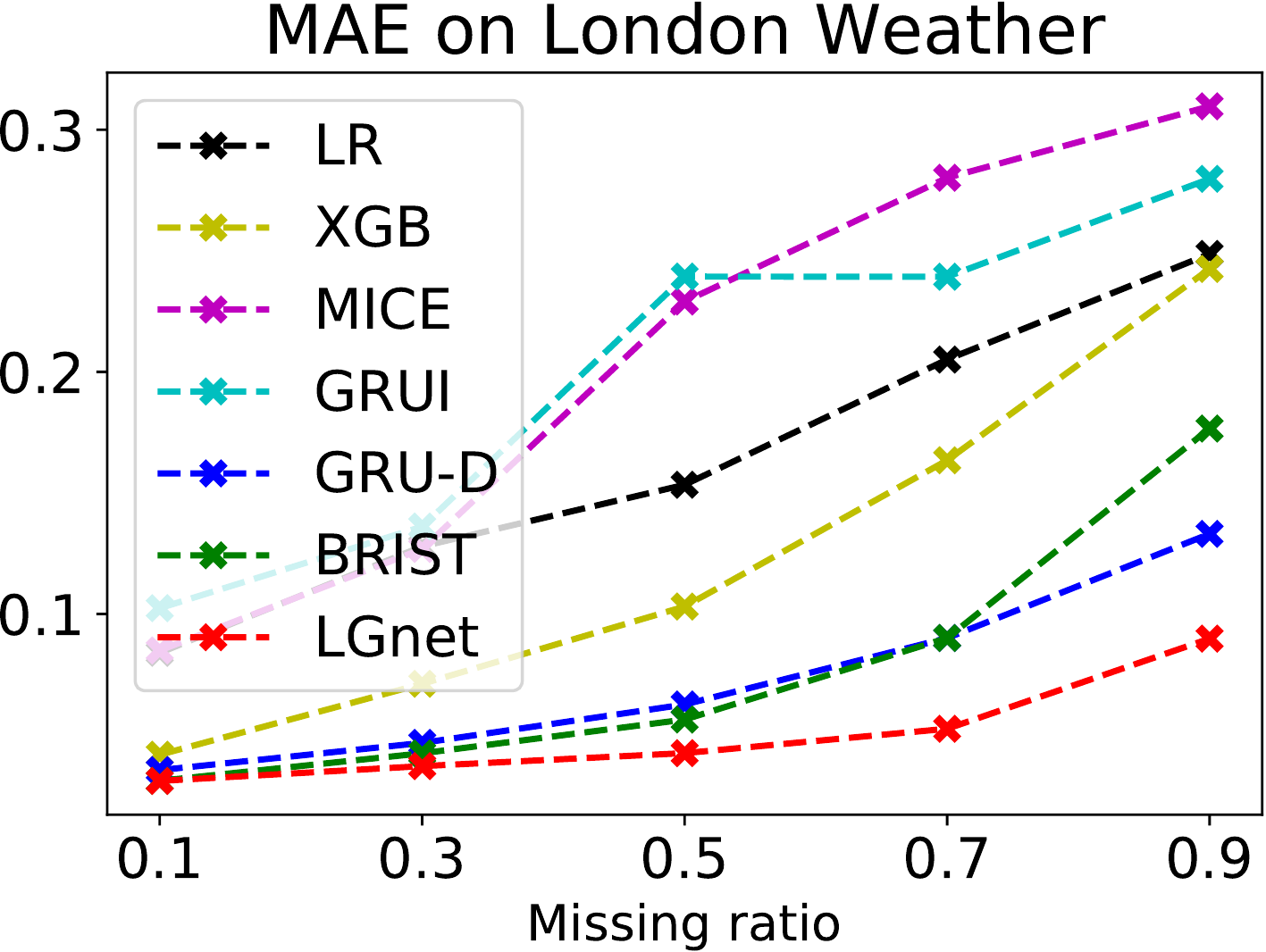}
        \end{subfigure}
        \caption{MTS forecasting performance on Porto taxi and London Weather with varying missing ratios.} 
        \label{ratio}
\end{figure}

Real-world applications could encounter various data missing conditions. It is interesting to understand the robustness of \ours under different missing ratios. To this end, we design experiments on two complete MTS datasets, including Porto taxi and London weather. In particular, for each dataset, we randomly drop $p$\% of observed values to generate synthetic missing condition and we alter $p$ from $\{10, 30, 50, 70, 90\}$. We train \ours and compared methods to forecast the next observation of all variables (i.e., $k=1$).
The performance of \ours and all compared methods in terms of RMSE and MAE varying $p$ is reported in Figure~\ref{ratio}.
Clearly, the forecasting error of non-RNN methods raises dramatically as $p$ increasing, because they fail to model  missing temporal patterns for the forecasting. GRU-D and BRITS explicitly handle missing values and achieve lower errors compared with LR and XGBoost. However, they fail to maintain accurate forecasting when the missing ratio is high. \ours achieves the highest accuracy even if the data is extremely sparse (e.g., $p= 90$), which illustrate the effectiveness of the memory module and the adversarial schema.
The global temporal patterns in memory module help \ours perform well as the missing ratio increasing. Extra guidance from the discriminator improves the capability of LSTM in modeling the global distribution of MTS.

\subsection{Ablation Study}
\subsubsection{Memory Module Analysis}
We analyze the contribution of the memory module. We create an ablation named $\text{\ours}_{adv}$ by removing the memory module from \ours, and use $\mathbf{q}_i$ as the input for LSTM. 
The performance of $\text{\ours}_{adv}$ is reported in Table \ref{remove_mem}. 
Obviously, \ours significantly out-performs $\text{\ours}_{adv}$, indicating that modeling global temporal dynamics with the memory module benefits the forecasting. 
Moreover, the performance improvement of \ours over $\text{\ours}_{adv}$ is relatively bigger as the missing ratio raises up. This is because local statistic features are less reliable with a high missing ratio. Under such circumstances, it is vital to leverage global patterns stored in the memory component as support to estimate missing values.

% We also visualize the attention score over the memory component to understand its behaviors, which confirms its ability to construct meaningful representation vector from a global perspective. Due to the page limit, we include detailed figures and analysis in the \textit{supplementary material}.

\begin{table}[]
\small
    \centering
\footnotesize
\caption{MTS forecasting performance of variants.}
\footnotesize
    \label{remove_mem}
    \resizebox{.95\columnwidth}{!}{
\begin{tabular}{c|ccc|ccc}
\hline \hline
\multirow{2}{*}{$p$} & \multicolumn{3}{c|}{Porto Taxi} & \multicolumn{3}{c}{London Weather} \\ \cline{2-7} 
                         & $\text{\ours}_{adv}$   & $\text{\ours}_{mem}$  & \ours                & $\text{\ours}_{adv}$   & $\text{\ours}_{mem}$   & \ours            \\ \hline
0.1                & 0.0157 & \textbf{0.0132}
& 0.0137           & 0.0627  & 0.0595     &  \textbf{0.0573} \\
0.3                    & 0.0241   & 0.0153 & \textbf{0.0151}   & 0.0773  & 0.0745     &  \textbf{0.0666} \\
0.5                 &  0.0303   & 0.0170  & \textbf{0.0168}  &    0.1020   & 0.0816  &  \textbf{0.0732} \\
0.7                   & 0.0451   & 0.0208  & \textbf{0.0204}   & 0.1287      & 0.0852 &  \textbf{0.0825} \\
0.9                   & 0.0710  & 0.0348   & \textbf{0.0337}      &  0.2015 & 0.1327   &  \textbf{0.1316} \\ \hline
\end{tabular}}
\end{table}

\subsubsection{Adversarial Schema Analysis}
We further study the effectiveness of adversarial training. The parameter $\lambda$ balances the weight between the forecasting loss and the adversarial part. A variant of \ours without the adversarial training (i.e., $\lambda=0$) is denoted as $\text{\ours}_{mem}$, and its performance is reported in Table \ref{remove_mem}.
Clearly, the adversarial training contributes a lot to the forecasting, reducing RMSE by 2\% -- 10\% under different circumstances. More concretely, more significant error reductions occur on London weather dataset compared with Proto taxi dataset. One possible reason is that MTS from London weather dataset contain more variables and have a better description of the realistic data distribution.
Besides, improvements from incorporating the discriminator are relatively greater when the missing ratio increases. This is because the original MTS forecasting objective is less efficient with a high missing ratio, as it only relies on observed parts of the time series. 
In conclusion, it is beneficial to introduce adversarial training for \ours.

\subsection{Hyper-parameter Analysis}
We investigate the sensitivity of $\lambda$, which balances the forecasting loss and the adversarial training part. Generally, More emphasis is put to the forecasting part when $\lambda$ is closer to 0. 
We alter the value of $\lambda$ among $\{0, 0.1, 1, 10, 100\}$ and report the performance of \ours on Beijing Air dataset with $k=1$. 
As shown in Table \ref{tab:lambda}, the forecasting accuracy of \ours first increases as $\lambda$ becomes larger. However, extremely large values of $\lambda$ result in low performances.

\begin{table}[t]
\centering
\footnotesize
\caption{Analysis of hyper-parameter $\lambda$.}
\resizebox{.95\columnwidth}{!}{
\begin{tabular}{c|ccccc}
\hline \hline
$\lambda$ & 0 & 0.1 & 1 & 10 & 100  \\ \hline
RMSE & 0.0483 & 0.0451 & 0.0471 & 0.0522 & 0.0518\\ 
MAPE & 0.0322 & 0.0300 & 0.0315 & 0.0360 & 0.0354 \\
\hline
\end{tabular}}
\label{tab:lambda}
\end{table}

\section{Conclusion} \label{sec:conclusion}
In this paper, we investigate a novel problem of exploring local and global temporal dynamics for MTS forecasting with missing values. We propose a new framework \ours, which adopts memory network to capture global temporal patterns using local statistics as keys. To make the generated MTS more realistic, we further adopt adversarial training to enhance the modeling of global temporal data distribution. Experimental results on four large-scale real-world datasets show the efficacy of \ours. %We also analyze the robustness of \ours under various missing ratios. Ablations are also designed to study individual contributions of the memory model and the adversarial training.

\section*{Acknowledgments}
This material is based upon work supported by, or in part by, the National Science Foundation (NSF) under grant \#1909702.

\fontsize{9.2pt}{10.2pt} \selectfont
\bibliographystyle{aaai}
\bibliography{ref.bib}

\begin{thebibliography}{}

\bibitem[\protect\citeauthoryear{Arjovsky, Chintala, and
  Bottou}{2017}]{arjovsky2017wasserstein}
Arjovsky, M.; Chintala, S.; and Bottou, L.
\newblock 2017.
\newblock Wasserstein gan.
\newblock {\em arXiv:1701.07875}.

\bibitem[\protect\citeauthoryear{Azur \bgroup et al\mbox.\egroup
  }{2011}]{azur2011multiple}
Azur, M.~J.; Stuart, E.~A.; Frangakis, C.; and Leaf, P.~J.
\newblock 2011.
\newblock Multiple imputation by chained equations: what is it and how does it
  work?
\newblock {\em International journal of methods in psychiatric research}
  20(1):40--49.

\bibitem[\protect\citeauthoryear{Bengio \bgroup et al\mbox.\egroup
  }{2015}]{bengio2015scheduled}
Bengio, S.; Vinyals, O.; Jaitly, N.; and Shazeer, N.
\newblock 2015.
\newblock Scheduled sampling for sequence prediction with recurrent neural
  networks.
\newblock In {\em NeurIPS},  1171--1179.

\bibitem[\protect\citeauthoryear{Box \bgroup et al\mbox.\egroup
  }{2015}]{box2015time}
Box, G.~E.; Jenkins, G.~M.; Reinsel, G.~C.; and Ljung, G.~M.
\newblock 2015.
\newblock {\em Time series analysis: forecasting and control}.
\newblock John Wiley \& Sons.

\bibitem[\protect\citeauthoryear{Cao \bgroup et al\mbox.\egroup
  }{2018}]{cao2018brits}
Cao, W.; Wang, D.; Li, J.; Zhou, H.; Li, L.; and Li, Y.
\newblock 2018.
\newblock Brits: Bidirectional recurrent imputation for time series.
\newblock {\em arXiv:1805.10572}.

\bibitem[\protect\citeauthoryear{Chang \bgroup et al\mbox.\egroup
  }{2018}]{chang2018memory}
Chang, Y.-Y.; Sun, F.-Y.; Wu, Y.-H.; and Lin, S.-D.
\newblock 2018.
\newblock A memory-network based solution for multivariate time-series
  forecasting.
\newblock {\em arXiv:1809.02105}.

\bibitem[\protect\citeauthoryear{Che \bgroup et al\mbox.\egroup
  }{2018}]{che2018recurrent}
Che, Z.; Purushotham, S.; Cho, K.; Sontag, D.; and Liu, Y.
\newblock 2018.
\newblock Recurrent neural networks for multivariate time series with missing
  values.
\newblock {\em Scientific reports} 8(1):6085.

\bibitem[\protect\citeauthoryear{Chen and Guestrin}{2016}]{chen2016xgboost}
Chen, T., and Guestrin, C.
\newblock 2016.
\newblock Xgboost: A scalable tree boosting system.
\newblock In {\em KDD},  785--794.
\newblock ACM.

\bibitem[\protect\citeauthoryear{Cho \bgroup et al\mbox.\egroup
  }{2014}]{cho2014learning}
Cho, K.; Van~Merri{\"e}nboer, B.; Gulcehre, C.; Bahdanau, D.; Bougares, F.;
  Schwenk, H.; and Bengio, Y.
\newblock 2014.
\newblock Learning phrase representations using rnn encoder-decoder for
  statistical machine translation.
\newblock {\em arXiv:1406.1078}.

\bibitem[\protect\citeauthoryear{Chung \bgroup et al\mbox.\egroup
  }{2014}]{chung2014empirical}
Chung, J.; Gulcehre, C.; Cho, K.; and Bengio, Y.
\newblock 2014.
\newblock Empirical evaluation of gated recurrent neural networks on sequence
  modeling.
\newblock {\em arXiv:1412.3555}.

\bibitem[\protect\citeauthoryear{Friedman, Hastie, and
  Tibshirani}{2001}]{friedman2001elements}
Friedman, J.; Hastie, T.; and Tibshirani, R.
\newblock 2001.
\newblock {\em The elements of statistical learning}, volume~1.
\newblock Springer series in statistics New York, NY, USA:.

\bibitem[\protect\citeauthoryear{Garc{\'\i}a-Laencina, Sancho-G{\'o}mez, and
  Figueiras-Vidal}{2010}]{garcia2010pattern}
Garc{\'\i}a-Laencina, P.~J.; Sancho-G{\'o}mez, J.-L.; and Figueiras-Vidal,
  A.~R.
\newblock 2010.
\newblock Pattern classification with missing data: a review.
\newblock {\em Neural Computing and Applications} 19(2):263--282.

\bibitem[\protect\citeauthoryear{Goodfellow \bgroup et al\mbox.\egroup
  }{2014}]{goodfellow2014generative}
Goodfellow, I.; Pouget-Abadie, J.; Mirza, M.; Xu, B.; Warde-Farley, D.; Ozair,
  S.; Courville, A.; and Bengio, Y.
\newblock 2014.
\newblock Generative adversarial nets.
\newblock In {\em NeurIPS},  2672--2680.

\bibitem[\protect\citeauthoryear{Hochreiter and
  Schmidhuber}{1997}]{hochreiter1997long}
Hochreiter, S., and Schmidhuber, J.
\newblock 1997.
\newblock Long short-term memory.
\newblock {\em Neural computation} 9(8):1735--1780.

\bibitem[\protect\citeauthoryear{King \bgroup et al\mbox.\egroup
  }{1998}]{king1998list}
King, G.; Honaker, J.; Joseph, A.; and Scheve, K.
\newblock 1998.
\newblock List-wise deletion is evil: what to do about missing data in
  political science.
\newblock In {\em APSA}.

\bibitem[\protect\citeauthoryear{Kumar \bgroup et al\mbox.\egroup
  }{2016}]{kumar2016ask}
Kumar, A.; Irsoy, O.; Ondruska, P.; Iyyer, M.; Bradbury, J.; Gulrajani, I.;
  Zhong, V.; Paulus, R.; and Socher, R.
\newblock 2016.
\newblock Ask me anything: Dynamic memory networks for natural language
  processing.
\newblock In {\em ICML},  1378--1387.

\bibitem[\protect\citeauthoryear{Lai \bgroup et al\mbox.\egroup
  }{2018}]{lai2018modeling}
Lai, G.; Chang, W.-C.; Yang, Y.; and Liu, H.
\newblock 2018.
\newblock Modeling long-and short-term temporal patterns with deep neural
  networks.
\newblock In {\em SIGIR}.
\newblock ACM.

\bibitem[\protect\citeauthoryear{Luo \bgroup et al\mbox.\egroup
  }{2018}]{luo2018multivariate}
Luo, Y.; Cai, X.; Zhang, Y.; Xu, J.; et~al.
\newblock 2018.
\newblock Multivariate time series imputation with generative adversarial
  networks.
\newblock In {\em NeurIPS},  1603--1614.

\bibitem[\protect\citeauthoryear{Marsh}{1998}]{marsh1998pairwise}
Marsh, H.~W.
\newblock 1998.
\newblock Pairwise deletion for missing data in structural equation models:
  Nonpositive definite matrices, parameter estimates, goodness of fit, and
  adjusted sample sizes.
\newblock {\em Structural Equation Modeling: A Multidisciplinary Journal} 5(1).

\bibitem[\protect\citeauthoryear{Nelwamondo, Mohamed, and
  Marwala}{2007}]{nelwamondo2007missing}
Nelwamondo, F.~V.; Mohamed, S.; and Marwala, T.
\newblock 2007.
\newblock Missing data: A comparison of neural network and expectation
  maximization techniques.
\newblock {\em Current Science}.

\bibitem[\protect\citeauthoryear{Qin \bgroup et al\mbox.\egroup
  }{2017}]{qin2017dual}
Qin, Y.; Song, D.; Chen, H.; Cheng, W.; Jiang, G.; and Cottrell, G.
\newblock 2017.
\newblock A dual-stage attention-based recurrent neural network for time series
  prediction.
\newblock {\em arXiv:1704.02971}.

\bibitem[\protect\citeauthoryear{Shu \bgroup et al\mbox.\egroup
  }{2018}]{shu2018deep}
Shu, K.; Wang, S.; Le, T.; Lee, D.; and Liu, H.
\newblock 2018.
\newblock Deep headline generation for clickbait detection.
\newblock In {\em ICDM}.
\newblock IEEE.

\bibitem[\protect\citeauthoryear{Silva \bgroup et al\mbox.\egroup
  }{2012}]{silva2012predicting}
Silva, I.; Moody, G.; Scott, D.~J.; Celi, L.~A.; and Mark, R.~G.
\newblock 2012.
\newblock Predicting in-hospital mortality of icu patients: The
  physionet/computing in cardiology challenge 2012.
\newblock {\em Computing in cardiology} 39:245.

\bibitem[\protect\citeauthoryear{Smola and
  Sch{\"o}lkopf}{2004}]{smola2004tutorial}
Smola, A.~J., and Sch{\"o}lkopf, B.
\newblock 2004.
\newblock A tutorial on support vector regression.
\newblock {\em Statistics and computing} 14(3):199--222.

\bibitem[\protect\citeauthoryear{Sukhbaatar \bgroup et al\mbox.\egroup
  }{2015}]{sukhbaatar2015end}
Sukhbaatar, S.; Weston, J.; Fergus, R.; et~al.
\newblock 2015.
\newblock End-to-end memory networks.
\newblock In {\em NeurIPS},  2440--2448.

\bibitem[\protect\citeauthoryear{Sun \bgroup et al\mbox.\egroup
  }{2019}]{sun2019megan}
Sun, Y.; Wang, S.; Hsieh, T.-Y.; Tang, X.; and Honavar, V.
\newblock 2019.
\newblock Megan: a generative adversarial network for multi-view network
  embedding.
\newblock {\em arXiv preprint arXiv:1909.01084}.

\bibitem[\protect\citeauthoryear{Tang \bgroup et al\mbox.\egroup
  }{2017}]{tang2017end}
Tang, J.; Wang, Y.; Zheng, K.; and Mei, Q.
\newblock 2017.
\newblock End-to-end learning for short text expansion.
\newblock In {\em KDD},  1105--1113.
\newblock ACM.

\bibitem[\protect\citeauthoryear{Tang \bgroup et al\mbox.\egroup
  }{2019}]{tang2019joint}
Tang, X.; Gong, B.; Yu, Y.; Yao, H.; Li, Y.; Xie, H.; and Wang, X.
\newblock 2019.
\newblock Joint modeling of dense and incomplete trajectories for citywide
  traffic volume inference.
\newblock In {\em WWW}.
\newblock ACM.

\bibitem[\protect\citeauthoryear{Wells \bgroup et al\mbox.\egroup
  }{2013}]{wells2013strategies}
Wells, B.~J.; Chagin, K.~M.; Nowacki, A.~S.; and Kattan, M.~W.
\newblock 2013.
\newblock Strategies for handling missing data in electronic health record
  derived data.
\newblock {\em Egems} 1(3).

\bibitem[\protect\citeauthoryear{Weston, Chopra, and
  Bordes}{2014}]{weston2014memory}
Weston, J.; Chopra, S.; and Bordes, A.
\newblock 2014.
\newblock Memory networks.
\newblock {\em arXiv:1410.3916}.

\bibitem[\protect\citeauthoryear{Wu \bgroup et al\mbox.\egroup
  }{2018}]{wu2018restful}
Wu, X.; Shi, B.; Dong, Y.; Huang, C.; Faust, L.; and Chawla, N.~V.
\newblock 2018.
\newblock Restful: Resolution-aware forecasting of behavioral time series data.
\newblock In {\em CIKM},  1073--1082.
\newblock ACM.

\bibitem[\protect\citeauthoryear{Wu \bgroup et al\mbox.\egroup
  }{2019}]{wu2019neural}
Wu, X.; Shi, B.; Dong, Y.; Huang, C.; and Chawla, N.~V.
\newblock 2019.
\newblock Neural tensor factorization for temporal interaction learning.
\newblock In {\em WSDM},  537--545.
\newblock ACM.

\bibitem[\protect\citeauthoryear{Xingjian \bgroup et al\mbox.\egroup
  }{2015}]{xingjian2015convolutional}
Xingjian, S.; Chen, Z.; Wang, H.; Yeung, D.-Y.; Wong, W.-K.; and Woo, W.-c.
\newblock 2015.
\newblock Convolutional lstm network: A machine learning approach for
  precipitation nowcasting.
\newblock In {\em NeurIPS},  802--810.

\bibitem[\protect\citeauthoryear{Yao \bgroup et al\mbox.\egroup
  }{2018}]{yao2018deep}
Yao, H.; Wu, F.; Ke, J.; Tang, X.; Jia, Y.; Lu, S.; Gong, P.; Ye, J.; and Li,
  Z.
\newblock 2018.
\newblock Deep multi-view spatial-temporal network for taxi demand prediction.
\newblock In {\em AAAI}.

\bibitem[\protect\citeauthoryear{Yao \bgroup et al\mbox.\egroup
  }{2019a}]{yao2019learning}
Yao, H.; Liu, Y.; Wei, Y.; Tang, X.; and Li, Z.
\newblock 2019a.
\newblock Learning from multiple cities: A meta-learning approach for
  spatial-temporal prediction.
\newblock {\em WWW}.

\bibitem[\protect\citeauthoryear{Yao \bgroup et al\mbox.\egroup
  }{2019b}]{yao2019revisiting}
Yao, H.; Tang, X.; Wei, H.; Zheng, G.; and Li, Z.
\newblock 2019b.
\newblock Revisiting spatial-temporal similarity: A deep learning framework for
  traffic prediction.
\newblock In {\em AAAI}.

\bibitem[\protect\citeauthoryear{Yi \bgroup et al\mbox.\egroup
  }{2016}]{yi2016st}
Yi, X.; Zheng, Y.; Zhang, J.; and Li, T.
\newblock 2016.
\newblock St-mvl: filling missing values in geo-sensory time series data.

\bibitem[\protect\citeauthoryear{Yoon, Zame, and van~der
  Schaar}{2017}]{Yoon2017MultidirectionalRN}
Yoon, J.; Zame, W.~R.; and van~der Schaar, M.
\newblock 2017.
\newblock Multi-directional recurrent neural networks : A novel method for
  estimating missing data.

\bibitem[\protect\citeauthoryear{Yu \bgroup et al\mbox.\egroup
  }{2017}]{yu2017seqgan}
Yu, L.; Zhang, W.; Wang, J.; and Yu, Y.
\newblock 2017.
\newblock Seqgan: Sequence generative adversarial nets with policy gradient.
\newblock In {\em AAAI}.

\end{thebibliography}

\end{document}